\documentclass{article}

\usepackage[preprint]{neurips_2026}

\usepackage[utf8]{inputenc} 
\usepackage[T1]{fontenc}    
\usepackage{hyperref}       
\usepackage{url}            
\usepackage{booktabs}       
\usepackage{amsfonts}       
\usepackage{nicefrac}       
\usepackage{microtype}      
\usepackage{xcolor}         
\usepackage{graphicx}
\usepackage{amsmath}
\title{Position: Vision-Language-Action Models Cannot Be Verified to Perform Physical Reasoning}

\author{
  Taozhao Chen \\
  School of Electrical and Computer Engineering\\
  The University of Sydney\\
  Sydney, NSW, Australia \\
  \texttt{tche8294@uni.sydney.edu.au} \\
  \And
  Ian Manchester \\
  School of Aerospace, Mechanical and Mechatronic Engineering\\
  The University of Sydney\\
  Sydney, NSW, Australia \\
  \texttt{ian.manchester@sydney.edu.au} \\
  \And
  Huaming Chen \\
  School of Electrical and Computer Engineering\\
  The University of Sydney\\
  Sydney, NSW, Australia \\
  \texttt{huaming.chen@sydney.edu.au} \\
}
\begin{document}

\maketitle

\begin{abstract}
Vision-Language-Action (VLA) systems, built on pretrained vision-language models (VLMs), have shown rapidly improving performance on robot manipulation benchmarks. These gains are commonly interpreted as evidence that semantic representations learned from internet-scale data transfer to physical execution generalization. \textbf{This position paper argues that the assumption underlying this interpretation—that semantic generalization is sufficient to support physical action decisions—has not been independently verified and cannot be tested under current evaluation protocols.}
We support this claim by decomposing VLA policies into semantic mapping and physical action decision, and showing that task success rate—the dominant evaluation metric—cannot distinguish between these two sources of capability. As a result, improvements in benchmark performance are consistent with multiple competing explanations, including semantic matching, distributional overlap, and genuine physical generalization. We further argue that this identifiability gap has been reinforced through \emph{narrative drift}, whereby successive systems inherit and strengthen prior interpretations of performance gains without isolating the underlying causal mechanism.
To address this limitation, we propose a research direction based on evaluation designs that introduce controlled variation to separately measure semantic and physical generalization. Such designs make it possible to causally attribute performance without requiring access to model internals, and to empirically assess the role of VLM backbones as semantic interfaces rather than implicit sources of physical competence. Our goal is not to refute the role of VLMs in robotics, but to clarify the conditions under which claims of physical generalization can be meaningfully evaluated.
\end{abstract}

\section{Introduction}

Teaching robots to perform manipulation tasks has long required extensive task-specific training data and engineering effort, with each new task demanding carefully curated demonstrations \cite{argall2009survey,ravichandar2020recent,zare2024survey} or hand-engineered reward functions \cite{kalashnikov2018scalable,sutton1998reinforcement}. 
The emergence of large vision-language models (VLMs)—pretrained on internet-scale visual and linguistic data—offers a compelling alternative: if such models encode broad semantic knowledge about objects, instructions, and goals, they may provide a generalizable foundation for robot control \cite{firoozi2025foundation,hu2023toward,kawaharazuka2025vision}.

\begin{figure}
    \centering
    \includegraphics[width=1\linewidth]{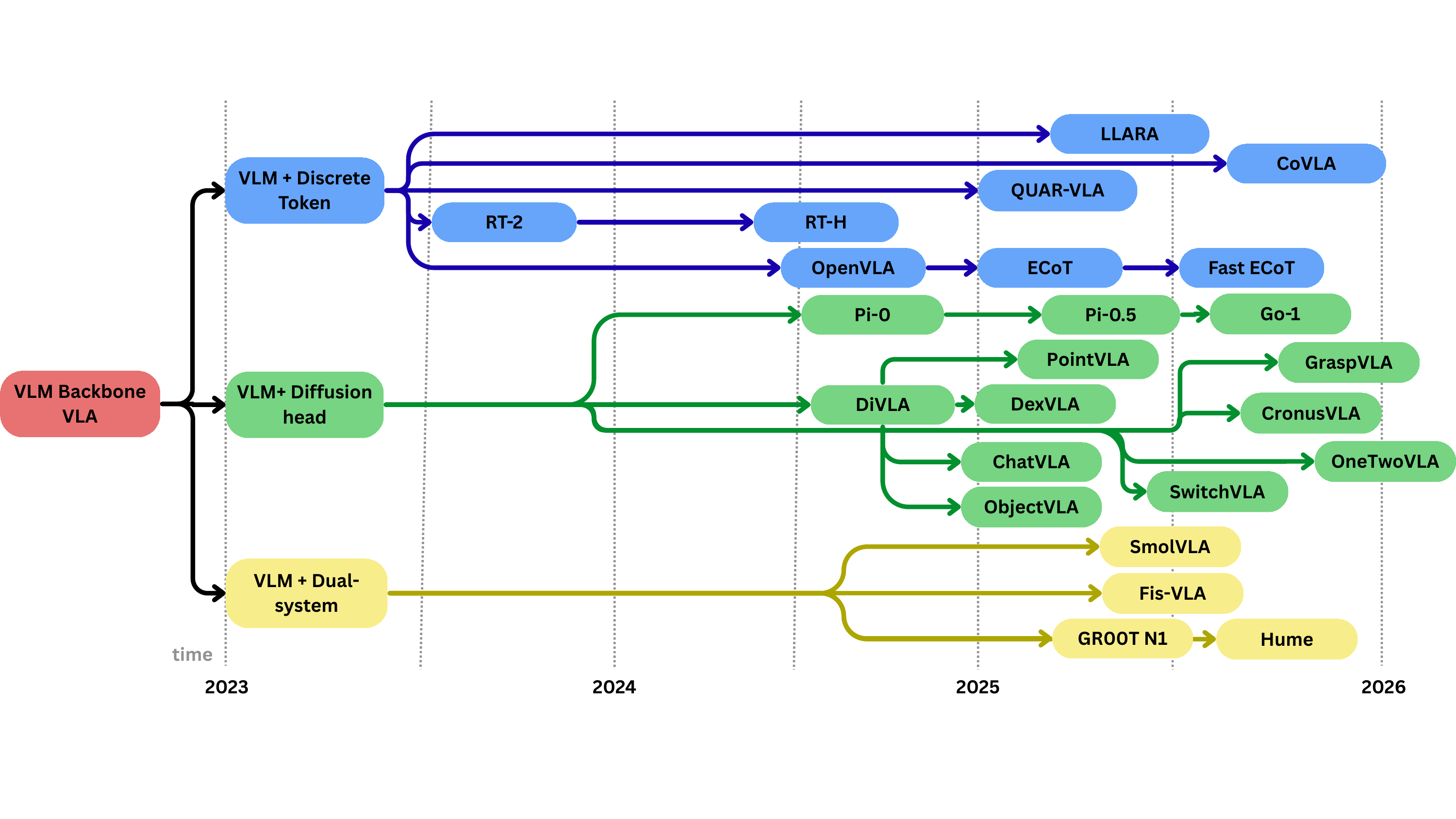}
    \caption{VLM-backbone VLA systems (2023–2026) organized into three architectural lineages: discrete token output, diffusion head, and dual-system. Arrows indicate architectural similarity only, not dependency or inheritance relationships.}
    \label{fig:placeholder}
\end{figure}

Since RT-2 demonstrated that pretrained VLMs can be repurposed as robot policies \cite{zitkovich2023rt}, the field has rapidly converged on the Vision-Language-Action (VLA) paradigm—end-to-end architectures that map visual observations and language instructions directly to action commands. As illustrated in Figure~\ref{fig:placeholder}, subsequent systems largely follow this formulation while exploring variations in output parameterization and control interfaces, forming several architectural lineages, including discrete token policies, diffusion-based policies, and dual-system designs. Despite these differences, they share a common structural core: a pretrained VLM backbone that mediates perception, language grounding, and action generation, and an implicit assumption—introduced by RT-2 and inherited by subsequent work—that semantic representations learned from large-scale vision-language pretraining are sufficient to guide physical action decisions. Under this premise, improvements in semantic generalization are expected to translate into more robust physical execution, motivating extensions along multiple axes, including diffusion-based action modeling, increased model scale, and additional modalities such as proprioception. This trajectory is reflected in systems such as OpenVLA \cite{kim2024openvla}, OpenVLA-OFT \cite{kim2025fine}, $\pi_0$ \cite{black2024pi_0}, and GR00T N1 \cite{bjorck2025gr00t}, which report steady benchmark improvements while maintaining this shared architectural and conceptual core.

However, task success measures outcomes, not the mechanism producing them. A central unresolved question is whether semantic generalization learned by VLMs transfers to physical execution generalization in robotic systems. These correspond to two distinct components: \emph{semantic generalization}, which maps observations and instructions to task-relevant abstractions, and \emph{physical execution}, which selects actions under constraints of dynamics, contact, embodiment, and temporal interaction.

\textbf{We argue that current VLA research relies on an implicit but unverified assumption: that semantic generalization is sufficient to support physical execution. This assumption has been reinforced through what we term narrative drift, whereby successive systems inherit and extend prior interpretations of performance gains without explicit causal verification.  Crucially, existing evaluation protocols do not provide the signals necessary to distinguish semantic generalization from physical execution, and therefore do not support causal attribution of performance improvements.}

We develop this argument as follows. Section~\ref{sec:assumption} formalizes the implicit assumption underlying VLM-backbone VLA research. Section~\ref{sec:argument} shows that this assumption is not identifiable under current evaluation protocols. Section~\ref{sec:consequences} analyzes the resulting systemic consequences. Section~\ref{sec:forward} proposes directions for evaluation reform. Section~\ref{sec:conclusion} concludes.

\section{The Implicit Assumption Underlying 
VLM-Backbone VLA Research}
\label{sec:assumption}

We formalize the distinction introduced in Section~1 by decomposing a VLA policy—mapping observations and instructions to actions—into two components. \textbf{Semantic mapping} transforms visual observations and language instructions into task-relevant representations (e.g., object categories, spatial relations, and intended goals), whereas \textbf{physical action decision} selects actions conditioned on this representation under constraints of dynamics, contact interactions, embodiment, and temporal execution. A capability is \emph{physical} if, holding the semantic interpretation fixed, variations in physical context (e.g., object mass, friction, geometry, or embodiment) require different actions for successful task completion. This distinction induces a criterion for attribution: improvements in task success can only be attributed to physical generalization if they remain robust under controlled variation of physical conditions that do not alter task semantics. It also implies distinct failure modes: semantic errors lead to incorrect task interpretation, while physical failures arise when the task is correctly identified but execution fails due to unmodeled physical factors—failure modes that are not distinguishable from task success rate alone.

We formalize this decomposition as:
\begin{equation}
    a_t \sim \pi\!\left(a_t \;\middle|\; z^{\text{sem}},\, z^{\text{phys}}\right)
    \label{eq:policy}
\end{equation}
where $z^{\text{sem}} = f_{\text{vlm}}(o_t, l)$ denotes the semantic representation extracted by the VLM backbone from current observation $o_t$ and language instruction $l$, and $z^{\text{phys}}$ denotes physical context—properties such as mass, friction, and contact geometry that are not recoverable from static image-text data. The \textbf{semantic sufficiency assumption} implicit in current VLA architectures is:
\begin{equation}
    \pi\!\left(a_t \mid z^{\text{sem}},\, z^{\text{phys}}\right) \approx \pi\!\left(a_t \mid z^{\text{sem}}\right)
    \label{eq:assumption}
\end{equation}

We decompose $z^{\text{sem}}$ into three constituent representations.
Categorical representations $z^{\text{cat}}$ encode object identity, category
membership, and visual appearance; they are acquired through large-scale
image--text contrastive training, in which visual and linguistic signals are
aligned via co-occurrence supervision~\cite{radford2021learning, li2023blip}.
Relational representations $z^{\text{rel}}$ encode spatial configurations,
affordance descriptions, and scene-level structure; they are learned through
grounding and captioning objectives that map linguistic relational expressions
to visual scene layouts~\cite{li2023blip, zeng2022socratic}.
Goal representations $z^{\text{goal}}$ encode task intent, sub-goal
decomposition, and success criteria; they are derived from instruction-following
and, in recent systems, chain-of-thought supervision over language and
image--text pairs~\cite{zitkovich2023rt, kim2024openvla}.
Each component is acquired from static image-text data: visual and linguistic
signals are jointly available, but action consequences are absent from the
supervision signal throughout~\cite{radford2021learning, li2023blip}.

This shared supervision structure defines the information boundary of
$z^{\text{sem}}$.
Physical properties such as mass distribution, surface friction, and contact
geometry are not recoverable from static visual observation: two objects may be
visually indistinguishable while differing substantially in physical
behavior~\cite{battaglia2013simulation, lerer2016learning}.
These properties are only revealed through interaction---through the consequences
of applied force, the resistance of contact, and the dynamics of
motion~\cite{agrawal2016learning, pinto2016supersizing}.
Because the training signal for $f_{\text{vlm}}$ contains no such interaction
outcomes, $z^{\text{sem}}$ cannot, in principle, encode $z^{\text{phys}}$
regardless of model scale or data volume~\cite{gibson2014ecological,
lake2017building}.
The semantic sufficiency assumption in Equation~\ref{eq:assumption} therefore
does not follow from the properties of VLM pretraining: it asserts that
$z^{\text{sem}}$ is sufficient for physical action decisions, but the
supervision structure of $f_{\text{vlm}}$ excludes precisely the signal that
would be required for this sufficiency to hold.

Unlike $z^{\text{sem}}$, which is explicitly computed via VLM encoding, current architectures provide no corresponding mechanism for acquiring $z^{\text{phys}}$, making \ref{eq:assumption} both structurally unjustified and empirically untested.

Despite substantial variation in output representations—discrete token policies trading action fidelity for semantic expressiveness \cite{zawalski2024robotic, duan2025fast, belkhale2024rt, li2024llara, arai2025covla, ding2024quar}, diffusion-based policies improving action continuity and precision \cite{zhou2025chatvla, zhu2025objectvla, bu2025agibot, li2026pointvla, li2025cronusvla, deng2025graspvla, lin2025onetwovla}, and dual-system architectures decoupling reasoning latency from execution speed \cite{chen2025fast, song2025hume, li2025switchvla}—the architectural core remains unchanged: all systems rely on a pretrained VLM backbone, with modifications concentrated in output heads and input modalities \cite{kim2024openvla, kim2025fine, black2024pi_0}. These design choices address execution limitations without isolating whether semantic representations are sufficient to support physical action decisions.

Nevertheless, capability claims have escalated alongside this unchanged core. Early work positioned VLMs as transferring web-scale semantic knowledge to robot control \cite{zitkovich2023rt}. Subsequent systems extended this to cross-task and cross-embodiment generalization \cite{kim2024openvla, kim2025fine}, while later work attributed semantic reasoning and problem-solving capabilities to VLM backbones in support of complex manipulation \cite{black2024pi_0}. More recent systems further suggest that scaling VLM backbones improves spatial reasoning and physical adaptation \cite{bjorck2025gr00t}. Across these claims, a component that has not fundamentally changed has been credited with progressively stronger capabilities.

Underlying this trajectory is a common but unverified condition: that improvements in semantic representations learned through VLM pretraining translate into robustness of physical action decisions. This condition is not necessarily stated explicitly in the cited works, but is required for their capability claims to hold under a causal interpretation. We refer to its progressive reinforcement across systems as \emph{narrative drift}.

\begin{table}[h]
\centering
\caption{Narrative drift in VLM-backbone VLA 
research: the architectural core and verification 
methodology remain unchanged while capability 
claims escalate continuously. These systems are selected as representative high-impact examples rather than an exhaustive survey.}
\label{tab:drift}
\scriptsize
\begin{tabular}{p{1.2cm}p{3.6cm}p{3.2cm}p{3cm}}
\toprule
\textbf{System} & 
\textbf{Primary Architectural Changes} & 
\textbf{Capability Claim} & 
\textbf{Assumption required for this claim to hold} \\
\midrule
RT-2
\cite{zitkovich2023rt} & 
Fine-tune PaLI-X/PaLM-E to output action 
tokens & 
Web-scale knowledge transfers to robot control & 
Semantic knowledge is sufficient to guide physical action selection \\
\midrule
OpenVLA
\cite{kim2024openvla} & 
Open-source VLM backbone; adjusted tokenization & 
Robust and generalizable visuomotor policies & 
Semantic generalization entails visuomotor generalization \\
\midrule
OpenVLA-OFT
\cite{kim2025fine} & 
Parallel decoding; multi-resolution input & 
Semantic generalization supports physical execution & 
Semantic competence transfers to execution via fine-tuning \\
\midrule
$\pi_0$
\cite{black2024pi_0} & 
Flow matching action expert & 
Inherits reasoning and problem-solving ability & 
Semantic reasoning transfers to manipulation competence \\
\midrule
GR00T N1
\cite{bjorck2025gr00t} & 
Dual-system architecture with proprioception & 
Stronger VLM improves adaptation in physical tasks & 
VLM strength reflects physical world modeling capability \\
\bottomrule
\end{tabular}
\end{table}

This pattern extends beyond the systems in Table~\ref{tab:drift} \cite{wen2024diffusion, zawalski2024robotic, arai2025covla, li2026semanticvla}. Across a broader range of VLM-backbone VLA works, improvements in task success rate—measured under largely unchanged evaluation protocols—are interpreted as evidence of increasingly strong physical execution generalization \cite{song2026reconvla, zhong2026dexgraspvla, Wang_2026, wang2026vlaadapter, zhang2026mole}. However, as introduced in Section~1, task success rate does not distinguish between semantic and physical contributions. Consequently, the same observed improvements are consistent with multiple explanations, including genuine physical generalization, improved semantic matching, or increased overlap between training and test distributions.

The consequence is not that the assumption is untestable in principle, but that it is not independently identifiable within the current methodological framework. Existing evidence does not permit attribution of performance gains to semantic or physical capabilities separately, and therefore cannot establish whether semantic representations are sufficient to support physical action decisions.
\section{The Structural Argument}\label{sec:argument}

\subsection{Semantic and Physical Representations Are Structurally Distinct}
\label{sec:argument:representations}

The assumption identified in Section~\ref{sec:assumption}—that semantic representations are sufficient to support physical action decisions—depends on whether the two representation types impose equivalent learning requirements. We argue that they do not.

Semantic representations encode what objects are and what instructions mean, including object identity, category membership, and relational descriptions grounded in language and visual appearance \cite{radford2021learning,li2023blip,zeng2022socratic}. Such representations can be acquired from large-scale image-text co-occurrence, where visual and linguistic signals are jointly available.

Physical representations encode how objects behave under interaction, including properties such as mass distribution, friction, contact geometry, and dynamic response to force \cite{billard2019trends}. These properties are not recoverable from static image-text data, as the consequences of action are absent from the training signal \cite{battaglia2013simulation,gibson2014ecological}. Increasing data scale does not resolve this limitation: supervision over physical state transitions is not present in the modality.

The two representation types therefore differ in their supervision requirements. Semantic representations can be learned from passive observation, whereas physical representations require interaction-based feedback over action outcomes \cite{pinto2016supersizing,agrawal2016learning}. A model trained exclusively on the former cannot be assumed to have acquired the latter.

This distinction implies that semantic generalization and physical execution generalization can dissociate. A system may correctly interpret the task while failing to execute it under novel physical conditions \cite{zitkovich2023rt,kim2025fine}. The assumption that semantic representations are sufficient for physical action decisions therefore does not follow from the properties of the training data or learning objective.
\subsection{Three Levels of Non-Identifiability in Current Evaluation}
\label{sec:benchmark}

Current benchmarks fail to support causal attribution at three distinct levels.
These levels are not independent: each deeper level is revealed only when the
shallower one is examined, and together they indicate that the evaluation gap
is more fundamental than it first appears.

\paragraph{Level 1: Attribution Non-Identifiability.}

Current robot manipulation benchmarks evaluate performance primarily through
task success rate \cite{kim2024openvla,liu2025eva,wang2026vla,bjorck2025gr00t}.
This metric does not distinguish between semantic and physical sources of
success or failure. Successful task completion may result from correct semantic
grounding, physically robust execution, or overlap between training and test
distributions. Failures may arise from incorrect task interpretation or from
inability to execute under novel physical conditions.

Staged evaluation protocols improve localization by decomposing tasks into
phases \cite{kim2025fine,nasiriany2024robocasa}, but do not resolve this
ambiguity. A failure at a specific stage remains consistent with both semantic
misinterpretation and physical execution error. These protocols identify where
failure occurs, but not why.

A critical limitation is that current benchmarks do not provide controlled
interventions that invalidate semantic matching while preserving task semantics.
As a result, policies that rely on distributional alignment between
instruction–observation pairs and action patterns cannot be distinguished from
those that model underlying physical dynamics. Attribution of performance to
semantic matching or physical generalization is therefore fundamentally
underdetermined.

However, even if attribution could be resolved, a deeper question remains:
\emph{where does the observed generalization come from?}

\paragraph{Level 2: Source Non-Identifiability.}

Even if attribution were resolved, the source of generalization would remain unidentifiable. Performance gains may arise from at least three sources—semantic priors from VLM pretraining, physical understanding from robot interaction data, or their interaction—yet current evaluation protocols control for none of them.

A concrete manifestation appears in the definition of out-of-distribution (OOD). OOD status is typically defined relative to robot demonstration data, not VLM pretraining data \cite{zitkovich2023rt,kim2024openvla,black2024pi_0}. Objects novel to the robot dataset may be familiar from pretraining, introducing an uncontrolled confound. Combined with task-specific fine-tuning, this creates a mismatch between nominal and actual sources of generalization.

As a result, observed improvements remain ambiguous in origin. Gains attributed to physical generalization may instead reflect prior semantic exposure or distributional overlap. Because current evaluation protocols neither control pretraining exposure nor isolate interaction-driven learning, they do not permit disentangling these sources.

This leads to source non-identifiability: the origin of generalization cannot be determined from observed outcomes.

\paragraph{Level 3: Representation-Level Non-Identifiability.}

Current benchmarks implicitly assume that all instructions are valid and executable within the agent’s physical context. Under this assumption, evaluation reduces to conditional action generation given a feasible task, and does not test whether the task itself should or can be executed.

This obscures the fate of capabilities inherited from pretrained vision-language backbones. Prior work shows that such models exhibit forms of constraint-aware reasoning and feasibility judgment in embodied settings—including scoring action affordances \cite{ahn2022can}, replanning after failures \cite{huang2022inner}, and recognizing infeasible instructions \cite{zhang2024recognizing}. However, it remains unclear whether these capabilities are preserved, degraded, or ignored during fine-tuning into action policies. This concern is supported by evidence that fine-tuning can overwrite pretrained representations \cite{zhou2025libero,yadav2025robust, fei2025libero,huang2026breaking}.

Because current benchmarks exclude infeasible, unsafe, or contradictory instructions, they provide no signal to detect the presence or absence of such capabilities. As a result, representation-level changes during policy learning remain unobservable.

This leads to representation-level non-identifiability: the effect of training on pre-existing capabilities cannot be measured, and potential degradation remains undetected.

Taken together, these three levels show that non-identifiability is not merely a measurement limitation but a structural property of the current evaluation paradigm. Attribution cannot be resolved at the level of outcomes, the source of generalization cannot be isolated across training signals, and changes in underlying representations cannot be observed during learning.

As a result, multiple latent factors—including semantic grounding, physical reasoning, and representation-level capabilities—are collapsed into a single observable metric. This prevents causal attribution and renders the central assumption of this paper—that semantic generalization is sufficient for physical execution—empirically untestable under current evaluation protocols.

\subsection{Empirical Signatures of Underdetermined Attribution}\label{sec:empirical}

The three levels of non-identifiability identified above are not merely
theoretical. If evaluation systematically cannot distinguish attribution,
source, and representation-level effects, the literature should exhibit a
predictable empirical signature: physical execution failures are observed but attributed to engineering factors, while the alternative explanation—failure of the semantic sufficiency assumption—is not evaluated.

Table~\ref{tab:signatures} illustrates this pattern across representative systems. In each case, observed limitations are consistent with failure of physical execution under novel conditions, yet are attributed to data coverage, model scale, or training procedure \cite{zitkovich2023rt,kim2024openvla,kim2025fine,black2024pi_0}. These attributions are plausible given the available evidence, but not uniquely supported by it.
\begin{table}[t]
\centering
\caption{Physical execution limitations across 
representative VLM-backbone VLA systems, and 
their attributed causes. In each case, an 
engineering attribution was adopted; the 
alternative---that the semantic sufficiency 
assumption has failed---was not considered.}
\label{tab:signatures}
\scriptsize
\begin{tabular}{p{1.4cm}p{6.8cm}p{3.8cm}}
\toprule
\textbf{System} & 
\textbf{Observed limitation} & 
\textbf{Author's attribution} \\
\midrule

RT-2~\cite{zitkovich2023rt} &
Model correctly identifies and approaches target 
objects but fails to control physical dynamics: 
pen rolling, banana center of mass displacement 
[App. G]. Manually identified through qualitative 
case analysis outside the primary evaluation 
protocol &
Physical skills are bounded by the robot training 
data distribution; more robot demonstration data 
needed \\

\midrule

RT-2~\cite{zitkovich2023rt} &
Larger VLM backbone parameters do not produce 
higher task success rates across evaluation 
conditions &
Model scale is not the bottleneck; 
data quality matters more than parameter count \\

\midrule

OpenVLA-OFT~\cite{kim2025fine} &
Policy relies on spurious visual correlations 
in multi-camera deployment; language grounding 
degrades despite unchanged task semantics 
[\S IV.C]. Distractor objects cause substantial 
success rate collapse while target 
localization remains intact &
Spurious correlations in training data; 
insufficient data coverage \\

\midrule

$\pi_0$~\cite{black2024pi_0} &
Task completion is unreliable across conditions; 
recovery from errors is fragile [\S VII]. 
Success rate varies substantially across tasks 
with no uniform failure pattern &
Data recipe problem; higher quality 
or more diverse data needed \\

\midrule
\bottomrule
\end{tabular}
\end{table}
This pattern admits two explanations. The first is that independent research groups consistently overlook the same alternative interpretation. The second is structural: if evaluation does not provide signals to distinguish semantic and physical failure, alternative explanations cannot be formulated or tested. The observed consistency of engineering attribution across systems supports the latter.

Under this interpretation, attribution in the literature reflects the limits of the evaluation framework rather than definitive evidence about underlying causes.

\section{Systemic Consequences}
\label{sec:consequences}

The semantic sufficiency assumption, when combined with an evaluation 
framework that does not provide identifiable signals for detecting its 
failure, produces three systemic consequences that compound over time.

\paragraph{Misallocation of Research Resources.}

When evaluation cannot attribute performance gains to their source, 
research investment decisions are made without reliable diagnostic signals 
\cite{sculley2015hidden,kiela2021dynabench}. 
The field cannot determine whether observed improvements reflect advances 
in semantic generalization, physical execution generalization, or 
distributional overlap—yet resource allocation depends precisely on this 
distinction \cite{d2022underspecification,geirhos2020shortcut}.

Investment in larger VLM backbones is justified if semantic capability 
transfers to physical execution; it is unjustified if the transfer does not 
occur and gains instead reflect distributional overlap. Likewise, investment 
in data collection depends on whether the bottleneck lies in semantic coverage 
or physical interaction diversity—two distinct requirements that existing 
benchmarks do not separately measure 
\cite{kalashnikov2018scalable,dasari2019robonet,nasiriany2024robocasa,mees2022calvin}.

This issue connects to a broader principle in the study of scaling. 
It is often argued that increasing scale alone is sufficient to achieve 
higher capability, with different methods primarily affecting efficiency 
rather than qualitative behavior 
\cite{hestness2017deep,kaplan2020scaling,huh2024platonic}. 
Under this view, evaluation plays a critical role: it determines whether 
scaling is applied to the correct underlying mechanism. If evaluation 
cannot distinguish between competing explanations of performance, scaling 
may amplify improvements along directions that do not correspond to the 
intended capability.

\paragraph{False Confidence in Deployment Readiness.}

Benchmark performance is the primary basis on which VLA systems are 
assessed for deployment readiness \cite{zitkovich2023rt,liu2023libero}. 
If benchmark success rates reflect distributional overlap rather than 
physical execution generalization, this confidence may not be warranted 
by the available evidence. OpenVLA-OFT reports near-perfect success rates 
on constrained tabletop tasks while the same system fails substantially 
when task-irrelevant distractor objects are introduced \cite{kim2025fine}: 
a perturbation that does not alter task semantics produces significant 
performance degradation.

This gap between benchmark performance and robustness to minimal 
perturbations is not reliably captured by current evaluation frameworks. 
A system can accumulate a strong benchmark record while its physical 
execution generalization remains unverified. In deployment contexts where 
physical configurations cannot be controlled, failure modes that benchmarks 
cannot detect become consequential: systems that succeed under training 
distribution overlap may fail when that overlap disappears, without prior 
evaluation signals that predict or characterize such failures.

\paragraph{Persistent Misdiagnosis of Architectural Failures.}

The attribution pattern identified in Section~\ref{sec:empirical}—physical 
execution limitations consistently attributed to engineering factors—is 
not merely a historical observation. It reflects a structural limitation 
of a field that lacks reliable tools to distinguish architectural causes 
from data or optimization effects \cite{d2022underspecification}.

When failures are attributed to insufficient data coverage, the proposed 
solution is more data. When failures are attributed to model capacity, 
the solution is larger models. When failures are attributed to action 
parameterization, the solution is alternative output representations. 
Each of these responses is reasonable under an engineering diagnosis. 
However, if the underlying cause is an unverified architectural assumption, 
these responses may leave the root issue unaddressed while continuing to 
consume resources.

The cost of persistent misdiagnosis is not limited to inefficiency. It 
produces a compounding divergence between the problems the field believes 
it is solving and the problems that must be solved for robust physical 
deployment. Systems may exhibit consistent improvements in benchmark 
performance while relying on mechanisms that do not generalize under 
physically novel conditions \cite{geirhos2020shortcut}. Without evaluation 
signals that support causal attribution, progress becomes difficult to 
distinguish from the accumulation of distributional alignment.

\section{Forward Path}
\label{sec:forward}

The consequences identified in Section~\ref{sec:consequences} arise from a 
single diagnostic limitation: the absence of identifiable signals for 
separately measuring semantic generalization and physical execution 
generalization. Addressing this limitation does not primarily require new 
model architectures; it requires evaluation designs that treat these two 
capabilities as distinct variables.

\paragraph{Evaluation Reform.}

Separating semantic and physical generalization requires experimental 
designs that independently manipulate semantic content and physical 
configuration. Two complementary interventions enable this separation.

To evaluate semantic generalization, physical conditions are held constant 
while semantic content varies. This can be achieved by altering linguistic 
descriptions of the same scene, or by introducing objects with visually 
similar appearance but different task-relevant properties. 

To evaluate physical execution generalization, semantic content is held 
constant while physical configuration varies. The same instruction and 
object category are presented under novel poses, surface conditions, or 
dynamics, requiring different action strategies for successful execution.

The key requirement is controlled variation: one factor must be held fixed 
while the other is systematically perturbed. Increasing dataset scale or 
benchmark diversity alone does not resolve the attribution problem if these 
factors remain entangled.

A particularly tractable instantiation of this design uses object properties 
that are linguistically expressible but not visually recoverable—such as 
mass, friction, or material composition 
\cite{zellers2019recognition,gibson1950perception,lerer2016learning}. 
A model that correctly grounds expressions such as ``the heavier cup'' or 
``the slippery object'' provides evidence of semantic generalization that 
cannot be reduced to visual pattern matching 
\cite{geirhos2020shortcut,brown2020language}. 

Evaluating whether this grounded representation leads to physically correct 
manipulation establishes a conditional test: semantic grounding is verified 
independently, and physical execution is evaluated given correct grounding 
\cite{agrawal2016learning}. This design introduces two observable outcomes—
semantic correctness and physical success—rather than collapsing both into 
a single task success signal. Importantly, this protocol does not require 
access to model internals; it relies only on observable behavior under 
controlled interventions \cite{10.1145/3442188.3445922}.

In addition, out-of-distribution status should be defined relative to all 
training data sources, including VLM pretraining corpora, rather than only 
robot demonstration datasets \cite{zitkovich2023rt,kim2024openvla,liu2023libero}. 
Without this distinction, nominally novel objects may remain familiar to the 
model, preventing attribution of performance gains to physical generalization.

\paragraph{Reorienting the Role of VLM Backbones.}

Recent work has highlighted a gap between high-level semantic reasoning and 
the requirements of physical execution, emphasizing the importance of 
three-dimensional structure, contact, and spatial relations in manipulation 
\cite{Chen_2026_ScienceRobotics,wang2023programmatically,cai2025spatialbot}. 
However, under current evaluation protocols, it remains unclear whether 
observed limitations arise from insufficient physical modeling or from 
confounding factors such as distributional overlap.

Evaluation designs that enable attribution have a direct implication for the 
role of VLM backbones. Once semantic and physical generalization can be 
measured separately, the contribution of VLM pretraining to each can be 
empirically assessed rather than assumed. This may reveal that VLM backbones 
provide strong semantic interfaces—grounding open-vocabulary instructions to 
visual targets \cite{li2023blip,radford2021learning,saycan2022arxiv}—while 
physical execution generalization depends on mechanisms not captured by 
passive image-text training \cite{lecun2022path,lake2017building}.

This perspective suggests a shift in architectural framing. Rather than 
treating VLM backbones as the foundation from which physical capability is 
expected to emerge, systems can be designed with explicitly separated 
components: a semantic interface for grounding language and perception, and 
a physical module for modeling action-conditioned dynamics 
\cite{wu2015galileo,tamkin2021understanding,raghu2019transfusion}. 
On such an architecture, the semantic sufficiency assumption need not be adopted \emph{from the outset}, because the two components are designed to serve distinct functions independently \cite{10.1145/3442188.3445922}.

This separation has precedent in developmental cognition. Infants exhibit 
sensitivity to physical properties such as object permanence and causal 
interaction prior to acquiring language 
\cite{lake2017building,leslie1987six}. Physical reasoning, on this account, 
develops as a distinct representational capacity rather than as a derivative 
of semantic knowledge.

This reframing does not prescribe a specific architecture. It shifts the 
design objective: from adapting semantic models to perform physical execution, 
to constructing systems in which physical execution generalization is 
measurable, attributable, and improvable independently of semantic grounding. 
Under such conditions, the contribution of VLM-based representations to 
physical performance becomes an empirical question rather than an assumption.

\section{Conclusion}\label{sec:conclusion}

This paper examined a foundational but unverified assumption underlying recent VLM-backbone VLA research: that semantic generalization learned from internet-scale data transfers to physical execution generalization in embodied systems. We showed that, under current evaluation protocols, this assumption is not independently identifiable. Task success rate—despite being the dominant metric—does not distinguish between semantic grounding and physical execution, and therefore does not support causal attribution of performance gains.

Reinterpreted through this lens, the steady improvement of benchmark results across VLA systems does not, by itself, establish progress in physical reasoning. Because semantic and physical factors remain entangled in both training and evaluation, the same empirical outcomes are consistent with multiple competing explanations, including improved semantic matching and distributional overlap. What appears as evidence of increasingly general physical capability may instead reflect the limits of what current benchmarks are able to measure.

The implication is methodological rather than architectural. Progress on physical execution generalization requires evaluation designs that introduce controlled variation, allowing semantic and physical contributions to be measured separately. This shift also reframes system design: instead of assuming that physical competence emerges from semantic representations, architectures can be structured to treat physical state modeling and semantic grounding as distinct components from the outset, enabling their roles to be empirically assessed rather than inferred.

More broadly, this analysis highlights a general principle for embodied AI: when evaluation does not provide identifiable signals for key capabilities, implicit assumptions can become entrenched as de facto premises of a field. 
This limitation has implications beyond attribution: it also constrains which capabilities are expressed and measured. Pretrained VLMs exhibit abilities—feasibility judgment, constraint awareness, context-sensitive refusal—that fine-tuning into action policies may degrade, yet task completion benchmarks provide no signal to detect such loss. 
Progress along the task completion axis alone does not guarantee movement toward general embodied intelligence. In this context, our contribution is not to refute the semantic sufficiency assumption, but to show that it has not yet been tested in a way that would allow it to be confirmed or rejected. Establishing such tests is a necessary step toward grounding claims of generalization in mechanisms that can be measured, attributed, and ultimately understood.
\bibliographystyle{plain}
\bibliography{ref}

@article{firoozi2025foundation,
  title={Foundation models in robotics: Applications, challenges, and the future},
  author={Firoozi, Roya and Tucker, Johnathan and Tian, Stephen and Majumdar, Anirudha and Sun, Jiankai and Liu, Weiyu and Zhu, Yuke and Song, Shuran and Kapoor, Ashish and Hausman, Karol and others},
  journal={The International Journal of Robotics Research},
  volume={44},
  number={5},
  pages={701--739},
  year={2025},
  publisher={Sage Publications Sage UK: London, England}
}

@article{kawaharazuka2025vision,
  title={Vision-language-action models for robotics: A review towards real-world applications},
  author={Kawaharazuka, Kento and Oh, Jihoon and Yamada, Jun and Posner, Ingmar and Zhu, Yuke},
  journal={IEEE Access},
  year={2025},
  publisher={IEEE}
}

@article{hu2023toward,
  title={Toward general-purpose robots via foundation models: A survey and meta-analysis},
  author={Hu, Yafei and Xie, Quanting and Jain, Vidhi and Francis, Jonathan and Patrikar, Jay and Keetha, Nikhil and Kim, Seungchan and Xie, Yaqi and Zhang, Tianyi and Fang, Hao-Shu and others},
  journal={arXiv preprint arXiv:2312.08782},
  year={2023}
}

@article{ravichandar2020recent,
  title={Recent advances in robot learning from demonstration},
  author={Ravichandar, Harish and Polydoros, Athanasios S and Chernova, Sonia and Billard, Aude},
  journal={Annual review of control, robotics, and autonomous systems},
  volume={3},
  number={1},
  pages={297--330},
  year={2020},
  publisher={Annual Reviews}
}

@article{argall2009survey,
  title={A survey of robot learning from demonstration},
  author={Argall, Brenna D and Chernova, Sonia and Veloso, Manuela and Browning, Brett},
  journal={Robotics and autonomous systems},
  volume={57},
  number={5},
  pages={469--483},
  year={2009},
  publisher={Elsevier}
}

@article{zare2024survey,
  title={A survey of imitation learning: Algorithms, recent developments, and challenges},
  author={Zare, Maryam and Kebria, Parham M and Khosravi, Abbas and Nahavandi, Saeid},
  journal={IEEE Transactions on Cybernetics},
  volume={54},
  number={12},
  pages={7173--7186},
  year={2024},
  publisher={IEEE}
}

@book{sutton1998reinforcement,
  title={Reinforcement learning: An introduction},
  author={Sutton, Richard S and Barto, Andrew G and others},
  volume={1},
  number={1},
  year={1998},
  publisher={MIT press Cambridge}
}

@inproceedings{kalashnikov2018scalable,
  title={Scalable deep reinforcement learning for vision-based robotic manipulation},
  author={Kalashnikov, Dmitry and Irpan, Alex and Pastor, Peter and Ibarz, Julian and Herzog, Alexander and Jang, Eric and Quillen, Deirdre and Holly, Ethan and Kalakrishnan, Mrinal and Vanhoucke, Vincent and others},
  booktitle={Conference on robot learning},
  pages={651--673},
  year={2018},
  organization={PMLR}
}

@inproceedings{zitkovich2023rt,
  title={Rt-2: Vision-language-action models transfer web knowledge to robotic control},
  author={Zitkovich, Brianna and Yu, Tianhe and Xu, Sichun and Xu, Peng and Xiao, Ted and Xia, Fei and Wu, Jialin and Wohlhart, Paul and Welker, Stefan and Wahid, Ayzaan and others},
  booktitle={Conference on Robot Learning},
  pages={2165--2183},
  year={2023},
  organization={PMLR}
}

@article{kim2024openvla,
  title={Openvla: An open-source vision-language-action model},
  author={Kim, Moo Jin and Pertsch, Karl and Karamcheti, Siddharth and Xiao, Ted and Balakrishna, Ashwin and Nair, Suraj and Rafailov, Rafael and Foster, Ethan and Lam, Grace and Sanketi, Pannag and others},
  journal={arXiv preprint arXiv:2406.09246},
  year={2024}
}

@article{kim2025fine,
  title={Fine-tuning vision-language-action models: Optimizing speed and success},
  author={Kim, Moo Jin and Finn, Chelsea and Liang, Percy},
  journal={arXiv preprint arXiv:2502.19645},
  year={2025}
}

@article{black2024pi_0,
  title={$pi\_0 $: A Vision-Language-Action Flow Model for General Robot Control},
  author={Black, Kevin and Brown, Noah and Driess, Danny and Esmail, Adnan and Equi, Michael and Finn, Chelsea and Fusai, Niccolo and Groom, Lachy and Hausman, Karol and Ichter, Brian and others},
  journal={arXiv preprint arXiv:2410.24164},
  year={2024}
}

@article{bjorck2025gr00t,
  title={Gr00t n1: An open foundation model for generalist humanoid robots},
  author={Bjorck, Johan and Casta{\~n}eda, Fernando and Cherniadev, Nikita and Da, Xingye and Ding, Runyu and Fan, Linxi and Fang, Yu and Fox, Dieter and Hu, Fengyuan and Huang, Spencer and others},
  journal={arXiv preprint arXiv:2503.14734},
  year={2025}
}

@inproceedings{radford2021learning,
  title={Learning transferable visual models from natural language supervision},
  author={Radford, Alec and Kim, Jong Wook and Hallacy, Chris and Ramesh, Aditya and Goh, Gabriel and Agarwal, Sandhini and Sastry, Girish and Askell, Amanda and Mishkin, Pamela and Clark, Jack and others},
  booktitle={International conference on machine learning},
  pages={8748--8763},
  year={2021},
  organization={PmLR}
}

@inproceedings{li2023blip,
  title={Blip-2: Bootstrapping language-image pre-training with frozen image encoders and large language models},
  author={Li, Junnan and Li, Dongxu and Savarese, Silvio and Hoi, Steven},
  booktitle={International conference on machine learning},
  pages={19730--19742},
  year={2023},
  organization={PMLR}
}

@article{zeng2022socratic,
  title={Socratic models: Composing zero-shot multimodal reasoning with language},
  author={Zeng, Andy and Attarian, Maria and Ichter, Brian and Choromanski, Krzysztof and Wong, Adrian and Welker, Stefan and Tombari, Federico and Purohit, Aveek and Ryoo, Michael and Sindhwani, Vikas and others},
  journal={arXiv preprint arXiv:2204.00598},
  year={2022}
}

@article{billard2019trends,
  title={Trends and challenges in robot manipulation},
  author={Billard, Aude and Kragic, Danica},
  journal={Science},
  volume={364},
  number={6446},
  pages={eaat8414},
  year={2019},
  publisher={American Association for the Advancement of Science}
}

@article{battaglia2013simulation,
  title={Simulation as an engine of physical scene understanding},
  author={Battaglia, Peter W and Hamrick, Jessica B and Tenenbaum, Joshua B},
  journal={Proceedings of the national academy of sciences},
  volume={110},
  number={45},
  pages={18327--18332},
  year={2013},
  publisher={National Academy of Sciences}
}

@book{gibson2014ecological,
  title={The ecological approach to visual perception: classic edition},
  author={Gibson, James J},
  year={2014},
  publisher={Psychology press}
}

@article{lake2017building,
  title={Building machines that learn and think like people},
  author={Lake, Brenden M and Ullman, Tomer D and Tenenbaum, Joshua B and Gershman, Samuel J},
  journal={Behavioral and brain sciences},
  volume={40},
  pages={e253},
  year={2017},
  publisher={Cambridge University Press}
}

@article{lecun2022path,
  title={A path towards autonomous machine intelligence version 0.9. 2, 2022-06-27},
  author={LeCun, Yann and others},
  journal={Open Review},
  volume={62},
  number={1},
  pages={1--62},
  year={2022}
}

@inproceedings{pinto2016supersizing,
  title={Supersizing self-supervision: Learning to grasp from 50k tries and 700 robot hours},
  author={Pinto, Lerrel and Gupta, Abhinav},
  booktitle={2016 IEEE international conference on robotics and automation (ICRA)},
  pages={3406--3413},
  year={2016},
  organization={IEEE}
}

@article{agrawal2016learning,
  title={Learning to poke by poking: Experiential learning of intuitive physics},
  author={Agrawal, Pulkit and Nair, Ashvin V and Abbeel, Pieter and Malik, Jitendra and Levine, Sergey},
  journal={Advances in neural information processing systems},
  volume={29},
  year={2016}
}

@article{liu2025eva,
  title={Eva-VLA: Evaluating Vision-Language-Action Models' Robustness Under Real-World Physical Variations},
  author={Liu, Hanqing and Ruan, Shouwei and Long, Jiahuan and Wu, Junqi and Hou, Jiacheng and Tang, Huili and Jiang, Tingsong and Zhou, Weien and Yao, Wen},
  journal={arXiv preprint arXiv:2509.18953},
  year={2025}
}

@article{wang2026vla,
  title={VLA Knows Its Limits},
  author={Wang, Haoxuan and Zhang, Gengyu and Yan, Yan and Kompella, Ramana Rao and Liu, Gaowen},
  journal={arXiv preprint arXiv:2602.21445},
  year={2026}
}

@article{nasiriany2024robocasa,
  title={Robocasa: Large-scale simulation of everyday tasks for generalist robots},
  author={Nasiriany, Soroush and Maddukuri, Abhiram and Zhang, Lance and Parikh, Adeet and Lo, Aaron and Joshi, Abhishek and Mandlekar, Ajay and Zhu, Yuke},
  journal={arXiv preprint arXiv:2406.02523},
  year={2024}
}

@article{liu2023libero,
  title={Libero: Benchmarking knowledge transfer for lifelong robot learning},
  author={Liu, Bo and Zhu, Yifeng and Gao, Chongkai and Feng, Yihao and Liu, Qiang and Zhu, Yuke and Stone, Peter},
  journal={Advances in Neural Information Processing Systems},
  volume={36},
  pages={44776--44791},
  year={2023}
}

@article{mees2022calvin,
  title={Calvin: A benchmark for language-conditioned policy learning for long-horizon robot manipulation tasks},
  author={Mees, Oier and Hermann, Lukas and Rosete-Beas, Erick and Burgard, Wolfram},
  journal={IEEE Robotics and Automation Letters},
  volume={7},
  number={3},
  pages={7327--7334},
  year={2022},
  publisher={IEEE}
}

@article{kaplan2020scaling,
  title={Scaling laws for neural language models},
  author={Kaplan, Jared and McCandlish, Sam and Henighan, Tom and Brown, Tom B and Chess, Benjamin and Child, Rewon and Gray, Scott and Radford, Alec and Wu, Jeffrey and Amodei, Dario},
  journal={arXiv preprint arXiv:2001.08361},
  year={2020}
}

@article{brown2020language,
  title={Language models are few-shot learners},
  author={Brown, Tom and Mann, Benjamin and Ryder, Nick and Subbiah, Melanie and Kaplan, Jared D and Dhariwal, Prafulla and Neelakantan, Arvind and Shyam, Pranav and Sastry, Girish and Askell, Amanda and others},
  journal={Advances in neural information processing systems},
  volume={33},
  pages={1877--1901},
  year={2020}
}

@article{sculley2015hidden,
  title={Hidden technical debt in machine learning systems},
  author={Sculley, David and Holt, Gary and Golovin, Daniel and Davydov, Eugene and Phillips, Todd and Ebner, Dietmar and Chaudhary, Vinay and Young, Michael and Crespo, Jean-Francois and Dennison, Dan},
  journal={Advances in neural information processing systems},
  volume={28},
  year={2015}
}

@inproceedings{kiela2021dynabench,
  title={Dynabench: Rethinking benchmarking in NLP},
  author={Kiela, Douwe and Bartolo, Max and Nie, Yixin and Kaushik, Divyansh and Geiger, Atticus and Wu, Zhengxuan and Vidgen, Bertie and Prasad, Grusha and Singh, Amanpreet and Ringshia, Pratik and others},
  booktitle={Proceedings of the 2021 conference of the North American chapter of the Association for Computational Linguistics: human language technologies},
  pages={4110--4124},
  year={2021}
}

@article{dasari2019robonet,
  title={Robonet: Large-scale multi-robot learning},
  author={Dasari, Sudeep and Ebert, Frederik and Tian, Stephen and Nair, Suraj and Bucher, Bernadette and Schmeckpeper, Karl and Singh, Siddharth and Levine, Sergey and Finn, Chelsea},
  journal={arXiv preprint arXiv:1910.11215},
  year={2019}
}

@article{d2022underspecification,
  title={Underspecification presents challenges for credibility in modern machine learning},
  author={D'Amour, Alexander and Heller, Katherine and Moldovan, Dan and Adlam, Ben and Alipanahi, Babak and Beutel, Alex and Chen, Christina and Deaton, Jonathan and Eisenstein, Jacob and Hoffman, Matthew D and others},
  journal={Journal of Machine Learning Research},
  volume={23},
  number={226},
  pages={1--61},
  year={2022}
}

@article{geirhos2020shortcut,
  title={Shortcut learning in deep neural networks},
  author={Geirhos, Robert and Jacobsen, J{\"o}rn-Henrik and Michaelis, Claudio and Zemel, Richard and Brendel, Wieland and Bethge, Matthias and Wichmann, Felix A},
  journal={Nature Machine Intelligence},
  volume={2},
  number={11},
  pages={665--673},
  year={2020},
  publisher={Nature Publishing Group UK London}
}

@inproceedings{zellers2019recognition,
  title={From recognition to cognition: Visual commonsense reasoning},
  author={Zellers, Rowan and Bisk, Yonatan and Farhadi, Ali and Choi, Yejin},
  booktitle={Proceedings of the IEEE/CVF conference on computer vision and pattern recognition},
  pages={6720--6731},
  year={2019}
}

@article{gibson1950perception,
  title={The perception of the visual world.},
  author={Gibson, James J},
  year={1950},
  publisher={Houghton Mifflin}
}

@inproceedings{lerer2016learning,
  title={Learning physical intuition of block towers by example},
  author={Lerer, Adam and Gross, Sam and Fergus, Rob},
  booktitle={International conference on machine learning},
  pages={430--438},
  year={2016},
  organization={PMLR}
}

@inproceedings{10.1145/3442188.3445922,
author = {Bender, Emily M. and Gebru, Timnit and McMillan-Major, Angelina and Shmitchell, Shmargaret},
title = {On the Dangers of Stochastic Parrots: Can Language Models Be Too Big?},
year = {2021},
isbn = {9781450383097},
publisher = {Association for Computing Machinery},
address = {New York, NY, USA},
url = {https://doi.org/10.1145/3442188.3445922},
doi = {10.1145/3442188.3445922},
booktitle = {Proceedings of the 2021 ACM Conference on Fairness, Accountability, and Transparency},
pages = {610–623},
numpages = {14},
location = {Virtual Event, Canada},
series = {FAccT '21}
}

@inproceedings{saycan2022arxiv,
    title={Do As I Can and Not As I Say: Grounding Language in Robotic Affordances},
    author={Michael Ahn and Anthony Brohan and Noah Brown and Yevgen Chebotar and Omar Cortes and Byron David and Chelsea Finn and Chuyuan Fu and Keerthana Gopalakrishnan and Karol Hausman and Alex Herzog and Daniel Ho and Jasmine Hsu and Julian Ibarz and Brian Ichter and Alex Irpan and Eric Jang and Rosario Jauregui Ruano and Kyle Jeffrey and Sally Jesmonth and Nikhil Joshi and Ryan Julian and Dmitry Kalashnikov and Yuheng Kuang and Kuang-Huei Lee and Sergey Levine and Yao Lu and Linda Luu and Carolina Parada and Peter Pastor and Jornell Quiambao and Kanishka Rao and Jarek Rettinghouse and Diego Reyes and Pierre Sermanet and Nicolas Sievers and Clayton Tan and Alexander Toshev and Vincent Vanhoucke and Fei Xia and Ted Xiao and Peng Xu and Sichun Xu and Mengyuan Yan and Andy Zeng},
    booktitle={arXiv preprint arXiv:2204.01691},
    year={2022}
}

@article{wu2015galileo,
  title={Galileo: Perceiving physical object properties by integrating a physics engine with deep learning},
  author={Wu, Jiajun and Yildirim, Ilker and Lim, Joseph J and Freeman, Bill and Tenenbaum, Josh},
  journal={Advances in neural information processing systems},
  volume={28},
  year={2015}
}

@article{tamkin2021understanding,
  title={Understanding the capabilities, limitations, and societal impact of large language models},
  author={Tamkin, Alex and Brundage, Miles and Clark, Jack and Ganguli, Deep},
  journal={arXiv preprint arXiv:2102.02503},
  year={2021}
}

@article{raghu2019transfusion,
  title={Transfusion: Understanding transfer learning for medical imaging},
  author={Raghu, Maithra and Zhang, Chiyuan and Kleinberg, Jon and Bengio, Samy},
  journal={Advances in neural information processing systems},
  volume={32},
  year={2019}
}

@article{leslie1987six,
  title={Do six-month-old infants perceive causality?},
  author={Leslie, Alan M and Keeble, Stephanie},
  journal={Cognition},
  volume={25},
  number={3},
  pages={265--288},
  year={1987},
  publisher={Elsevier}
}

@article{zawalski2024robotic,
  title={Robotic control via embodied chain-of-thought reasoning},
  author={Zawalski, Micha{\l} and Chen, William and Pertsch, Karl and Mees, Oier and Finn, Chelsea and Levine, Sergey},
  journal={arXiv preprint arXiv:2407.08693},
  year={2024}
}

@article{duan2025fast,
  title={Fast ecot: Efficient embodied chain-of-thought via thoughts reuse},
  author={Duan, Zhekai and Zhang, Yuan and Geng, Shikai and Liu, Gaowen and Boedecker, Joschka and Lu, Chris Xiaoxuan},
  journal={arXiv preprint arXiv:2506.07639},
  year={2025}
}

@article{belkhale2024rt,
  title={Rt-h: Action hierarchies using language},
  author={Belkhale, Suneel and Ding, Tianli and Xiao, Ted and Sermanet, Pierre and Vuong, Quon and Tompson, Jonathan and Chebotar, Yevgen and Dwibedi, Debidatta and Sadigh, Dorsa},
  journal={arXiv preprint arXiv:2403.01823},
  year={2024}
}

@article{li2024llara,
  title={Llara: Supercharging robot learning data for vision-language policy},
  author={Li, Xiang and Mata, Cristina and Park, Jongwoo and Kahatapitiya, Kumara and Jang, Yoo Sung and Shang, Jinghuan and Ranasinghe, Kanchana and Burgert, Ryan and Cai, Mu and Lee, Yong Jae and others},
  journal={arXiv preprint arXiv:2406.20095},
  year={2024}
}

@inproceedings{arai2025covla,
  title={Covla: Comprehensive vision-language-action dataset for autonomous driving},
  author={Arai, Hidehisa and Miwa, Keita and Sasaki, Kento and Watanabe, Kohei and Yamaguchi, Yu and Aoki, Shunsuke and Yamamoto, Issei},
  booktitle={2025 IEEE/CVF Winter Conference on Applications of Computer Vision (WACV)},
  pages={1933--1943},
  year={2025},
  organization={IEEE}
}

@inproceedings{ding2024quar,
  title={Quar-vla: Vision-language-action model for quadruped robots},
  author={Ding, Pengxiang and Zhao, Han and Zhang, Wenjie and Song, Wenxuan and Zhang, Min and Huang, Siteng and Yang, Ningxi and Wang, Donglin},
  booktitle={European Conference on Computer Vision},
  pages={352--367},
  year={2024},
  organization={Springer}
}

@article{wen2024diffusion,
  title={Diffusion-vla: Generalizable and interpretable robot foundation model via self-generated reasoning},
  author={Wen, Junjie and Zhu, Minjie and Zhu, Yichen and Tang, Zhibin and Li, Jinming and Zhou, Zhongyi and Li, Chengmeng and Liu, Xiaoyu and Peng, Yaxin and Shen, Chaomin and others},
  journal={arXiv preprint arXiv:2412.03293},
  year={2024}
}

@inproceedings{zhou2025chatvla,
  title={Chatvla: Unified multimodal understanding and robot control with vision-language-action model},
  author={Zhou, Zhongyi and Zhu, Yichen and Zhu, Minjie and Wen, Junjie and Liu, Ning and Xu, Zhiyuan and Meng, Weibin and Peng, Yaxin and Shen, Chaomin and Feng, Feifei and others},
  booktitle={Proceedings of the 2025 Conference on Empirical Methods in Natural Language Processing},
  pages={5377--5395},
  year={2025}
}

@article{zhu2025objectvla,
  title={Objectvla: End-to-end open-world object manipulation without demonstration},
  author={Zhu, Minjie and Zhu, Yichen and Li, Jinming and Zhou, Zhongyi and Wen, Junjie and Liu, Xiaoyu and Shen, Chaomin and Peng, Yaxin and Feng, Feifei},
  journal={arXiv preprint arXiv:2502.19250},
  year={2025}
}

@article{bu2025agibot,
  title={Agibot world colosseo: A large-scale manipulation platform for scalable and intelligent embodied systems},
  author={Bu, Qingwen and Cai, Jisong and Chen, Li and Cui, Xiuqi and Ding, Yan and Feng, Siyuan and Gao, Shenyuan and He, Xindong and Hu, Xuan and Huang, Xu and others},
  journal={arXiv preprint arXiv:2503.06669},
  year={2025}
}

@article{li2026pointvla,
  title={Pointvla: Injecting the 3d world into vision-language-action models},
  author={Li, Chengmeng and Wen, Junjie and Peng, Yaxin and Peng, Yan and Zhu, Yichen},
  journal={IEEE Robotics and Automation Letters},
  volume={11},
  number={3},
  pages={2506--2513},
  year={2026},
  publisher={IEEE}
}

@article{chen2025fast,
  title={Fast-in-slow: A dual-system foundation model unifying fast manipulation within slow reasoning},
  author={Chen, Hao and Liu, Jiaming and Gu, Chenyang and Liu, Zhuoyang and Zhang, Renrui and Li, Xiaoqi and He, Xiao and Guo, Yandong and Fu, Chi-Wing and Zhang, Shanghang and others},
  journal={arXiv preprint arXiv:2506.01953},
  year={2025}
}

@article{li2025cronusvla,
  title={CronusVLA: Towards Efficient and Robust Manipulation via Multi-Frame Vision-Language-Action Modeling},
  author={Li, Hao and Yang, Shuai and Chen, Yilun and Chen, Xinyi and Yang, Xiaoda and Tian, Yang and Wang, Hanqing and Wang, Tai and Lin, Dahua and Zhao, Feng and others},
  journal={arXiv preprint arXiv:2506.19816},
  year={2025}
}

@article{deng2025graspvla,
  title={Graspvla: a grasping foundation model pre-trained on billion-scale synthetic action data},
  author={Deng, Shengliang and Yan, Mi and Wei, Songlin and Ma, Haixin and Yang, Yuxin and Chen, Jiayi and Zhang, Zhiqi and Yang, Taoyu and Zhang, Xuheng and Zhang, Wenhao and others},
  journal={arXiv preprint arXiv:2505.03233},
  year={2025}
}

@article{lin2025onetwovla,
  title={Onetwovla: A unified vision-language-action model with adaptive reasoning},
  author={Lin, Fanqi and Nai, Ruiqian and Hu, Yingdong and You, Jiacheng and Zhao, Junming and Gao, Yang},
  journal={arXiv preprint arXiv:2505.11917},
  year={2025}
}

@article{song2025hume,
  title={Hume: Introducing system-2 thinking in visual-language-action model},
  author={Song, Haoming and Qu, Delin and Yao, Yuanqi and Chen, Qizhi and Lv, Qi and Tang, Yiwen and Shi, Modi and Ren, Guanghui and Yao, Maoqing and Zhao, Bin and others},
  journal={arXiv preprint arXiv:2505.21432},
  year={2025}
}

@article{li2025switchvla,
  title={Switchvla: Execution-aware task switching for vision-language-action models},
  author={Li, Meng and Zhao, Zhen and Che, Zhengping and Liao, Fei and Wu, Kun and Xu, Zhiyuan and Ren, Pei and Jin, Zhao and Liu, Ning and Tang, Jian},
  journal={arXiv preprint arXiv:2506.03574},
  year={2025}
}

@inproceedings{li2026semanticvla,
  title={Semanticvla: Semantic-aligned sparsification and enhancement for efficient robotic manipulation},
  author={Li, Wei and Zhang, Renshan and Shao, Rui and Fang, Zhijian and Zhou, Kaiwen and Tian, Zhuotao and Nie, Liqiang},
  booktitle={Proceedings of the AAAI Conference on Artificial Intelligence},
  volume={40},
  number={22},
  pages={18397--18405},
  year={2026}
}

@inproceedings{song2026reconvla,
  title={Reconvla: Reconstructive vision-language-action model as effective robot perceiver},
  author={Song, Wenxuan and Zhou, Ziyang and Zhao, Han and Chen, Jiayi and Ding, Pengxiang and Yan, Haodong and Huang, Yuxin and Tang, Feilong and Wang, Donglin and Li, Haoang},
  booktitle={Proceedings of the AAAI Conference on Artificial Intelligence},
  volume={40},
  number={22},
  pages={18549--18557},
  year={2026}
}

@inproceedings{zhong2026dexgraspvla,
  title={Dexgraspvla: A vision-language-action framework towards general dexterous grasping},
  author={Zhong, Yifan and Huang, Xuchuan and Li, Ruochong and Zhang, Ceyao and Chen, Zhang and Guan, Tianrui and Zeng, Fanlian and Lui, Ka Nam and Ye, Yuyao and Liang, Yitao and others},
  booktitle={Proceedings of the AAAI Conference on Artificial Intelligence},
  volume={40},
  number={22},
  pages={18836--18844},
  year={2026}
}

@article{Wang_2026,
  author  = {Wang, Junming},
  title   = {LatentVLA: Taming Latent Space for Generalizable and Long-Horizon Bimanual Manipulation},
  journal = {Proceedings of the AAAI Conference on Artificial Intelligence},
  year    = {2026},
  volume  = {40},
  number  = {22},
  pages   = {18593-18601},
  month   = {Mar.},
  doi     = {10.1609/aaai.v40i22.38926},
  url     = {https://ojs.aaai.org/index.php/AAAI/article/view/38926}
}

@inproceedings{wang2026vlaadapter,
  title={Vla-adapter: An effective paradigm for tiny-scale vision-language-action model},
  author={Wang, Yihao and Ding, Pengxiang and Li, Lingxiao and Cui, Can and Ge, Zirui and Tong, Xinyang and Song, Wenxuan and Zhao, Han and Zhao, Wei and Hou, Pengxu and others},
  booktitle={Proceedings of the AAAI conference on artificial intelligence},
  volume={40},
  number={22},
  pages={18638--18646},
  year={2026}
}

@inproceedings{zhang2026mole,
  title={Mole-vla: Dynamic layer-skipping vision language action model via mixture-of-layers for efficient robot manipulation},
  author={Zhang, Rongyu and Dong, Menghang and Zhang, Yuan and Heng, Liang and Chi, Xiaowei and Dai, Gaole and Du, Li and Wang, Dan and Du, Yuan and Zhang, Shanghang},
  booktitle={Proceedings of the AAAI Conference on Artificial Intelligence},
  volume={40},
  number={22},
  pages={18764--18772},
  year={2026}
}

@article{hestness2017deep,
  title={Deep learning scaling is predictable, empirically},
  author={Hestness, Joel and Narang, Sharan and Ardalani, Newsha and Diamos, Gregory and Jun, Heewoo and Kianinejad, Hassan and Patwary, Md Mostofa Ali and Yang, Yang and Zhou, Yanqi},
  journal={arXiv preprint arXiv:1712.00409},
  year={2017}
}

@article{huh2024platonic,
  title={The platonic representation hypothesis},
  author={Huh, Minyoung and Cheung, Brian and Wang, Tongzhou and Isola, Phillip},
  journal={arXiv preprint arXiv:2405.07987},
  year={2024}
}

@article{Chen_2026_ScienceRobotics,
  author  = {Kai Chen and Chengkun Li and Chang Tu and Jiahui Pan and Yiyao Ma and Wei Chen and Zhongxiang Zhou and Xuecheng Xu and Stephen James and Chi-Wing Fu and Rong Xiong and Pieter Abbeel and Yun-Hui Liu and Qi Dou},
  title   = {A retrieval-augmented framework enabling VLM spatial awareness for object-centric robot manipulation},
  journal = {Science Robotics},
  year    = {2026},
  volume  = {11},
  number  = {113},
  pages   = {eaea2092},
  doi     = {10.1126/scirobotics.aea2092},
  url     = {https://www.science.org/doi/abs/10.1126/scirobotics.aea2092}
}

@article{wang2023programmatically,
  title={Programmatically grounded, compositionally generalizable robotic manipulation},
  author={Wang, Renhao and Mao, Jiayuan and Hsu, Joy and Zhao, Hang and Wu, Jiajun and Gao, Yang},
  journal={arXiv preprint arXiv:2304.13826},
  year={2023}
}

@inproceedings{cai2025spatialbot,
  title={Spatialbot: Precise spatial understanding with vision language models},
  author={Cai, Wenxiao and Ponomarenko, Iaroslav and Yuan, Jianhao and Li, Xiaoqi and Yang, Wankou and Dong, Hao and Zhao, Bo},
  booktitle={2025 IEEE International Conference on Robotics and Automation (ICRA)},
  pages={9490--9498},
  year={2025},
  organization={IEEE}
}

@article{huang2026breaking,
  title={Breaking Lock-In: Preserving Steerability under Low-Data VLA Post-Training},
  author={Huang, Suning and Shao, Jiaqi and Wang, Ke and Chen, Qianzhong and Sun, Jiankai and Guo, Yanjiang and Schwager, Mac and Bohg, Jeannette},
  journal={arXiv preprint arXiv:2604.23121},
  year={2026}
}

@article{yadav2025robust,
  title={Robust Finetuning of Vision-Language-Action Robot Policies via Parameter Merging},
  author={Yadav, Yajat and Zhou, Zhiyuan and Wagenmaker, Andrew and Pertsch, Karl and Levine, Sergey},
  journal={arXiv preprint arXiv:2512.08333},
  year={2025}
}

@article{zhou2025libero,
  title={LIBERO-PRO: Towards Robust and Fair Evaluation of Vision-Language-Action Models Beyond Memorization},
  author={Zhou, Xueyang and Xu, Yangming and Tie, Guiyao and Chen, Yongchao and Zhang, Guowen and Chu, Duanfeng and Zhou, Pan and Sun, Lichao},
  journal={arXiv preprint arXiv:2510.03827},
  year={2025}
}

@article{fei2025libero,
  title={Libero-plus: In-depth robustness analysis of vision-language-action models},
  author={Fei, Senyu and Wang, Siyin and Shi, Junhao and Dai, Zihao and Cai, Jikun and Qian, Pengfang and Ji, Li and He, Xinzhe and Zhang, Shiduo and Fei, Zhaoye and others},
  journal={arXiv preprint arXiv:2510.13626},
  year={2025}
}

@article{ahn2022can,
  title={Do as i can, not as i say: Grounding language in robotic affordances},
  author={Ahn, Michael and Brohan, Anthony and Brown, Noah and Chebotar, Yevgen and Cortes, Omar and David, Byron and Finn, Chelsea and Fu, Chuyuan and Gopalakrishnan, Keerthana and Hausman, Karol and others},
  journal={arXiv preprint arXiv:2204.01691},
  year={2022}
}

@article{huang2022inner,
  title={Inner monologue: Embodied reasoning through planning with language models},
  author={Huang, Wenlong and Xia, Fei and Xiao, Ted and Chan, Harris and Liang, Jacky and Florence, Pete and Zeng, Andy and Tompson, Jonathan and Mordatch, Igor and Chebotar, Yevgen and others},
  journal={arXiv preprint arXiv:2207.05608},
  year={2022}
}

@article{zhang2024recognizing,
  title={Recognizing Limits: Investigating Infeasibility in Large Language Models},
  author={Zhang, Wenbo and Xu, Zihang and Cai, Hengrui},
  journal={arXiv preprint arXiv:2408.05873},
  year={2024}
}


\end{document}